\documentclass{article}
\usepackage{spconf,amsmath,amssymb,graphicx,booktabs}
\usepackage{iftex}
\ifPDFTeX
  \usepackage[T1]{fontenc} 
\else
  \usepackage{fontspec}
\fi
\usepackage{xcolor}

\newcommand{\cdashline}[1]{\noalign{\nointerlineskip\hbox to\linewidth{\leaders\hbox to 3pt{\hss\vrule height 0.4pt depth 0pt width 1.5pt\hss}\hfill}}}
\newcommand{\best}[1]{\textcolor{red!85!black}{\textbf{#1}}}    
\newcommand{\second}[1]{\textcolor{blue!80!black}{\underline{#1}}}

\usepackage{enumitem}
\usepackage[hidelinks]{hyperref} 
\setlist{nosep, leftmargin=14pt}

\title{Dropping Tokens Without Dropping Them: Causal Attribution for Diabetic Retinopathy}
\name{Pramath Girish Haritz \qquad Chandra Sekhar Seelamantula%
  \thanks{\small\textbf{Funding support:} We gratefully acknowledge funding support from ZEISS India and the Kotak-IISc AI/ML Centre. The funders had no role in the study design, data analysis, interpretation of the results, or the decision to publish.\\
  \textbf{AI disclosure statement:} Claude (Anthropic) was used to assist with the experiments and to polish the presentation. The authors take full responsibility for the work.}} 
\address{Indian Institute of Science}

\begin{document}
\ninept
\maketitle

\begin{abstract}
In this work, we investigate whether attention-based attribution maps in a vision transformer provide causal explanations for diabetic retinopathy (DR) classification. We introduce a principled decomposition of attention into two attribution maps: the contribution ($c_t$), which satisfies the completeness axiom in explainable artificial intelligence (xAI), and the pull ($p_t$), which quantifies causal explanations. We define an identity relating $c_t$ and $p_t$ that characterizes when $c_t$ fails to capture causality. We test whether each map is causally faithful on the MAPLES-DR dataset by dropping/inserting tokens, and report Necessity ($N$), Sufficiency ($Su$), and Faithfulness Correlation ($FC$). The pull $p_t$ achieves the highest causal alignment ($FC = 0.53$, $N = 0.71\%$, $Su = 3.0\%$), outperforming Attention Rollout, Grad-CAM, and AttnLRP. Occlusion retains better sufficiency ($Su = 1.6\%$) but is about 800 times slower. Furthermore, on MAPLES-DR, $p_t$ achieves $70\%$ of the average precision (AP) of HRDecoder, a state-of-the-art supervised lesion segmentation model, despite the absence of pixel-level supervision. Together, these findings show that the pull $p_t$ and the contribution $c_t$ form computationally efficient attribution maps for reliable causal interpretation in safety-critical medical diagnostics. The code and demo are available at \url{https://github.com/pramath-haritz/dr-xai}.
\end{abstract}


\section{Introduction}
\label{sec:intro}

\begin{figure*}[t]
\centering
\includegraphics[width=\textwidth]{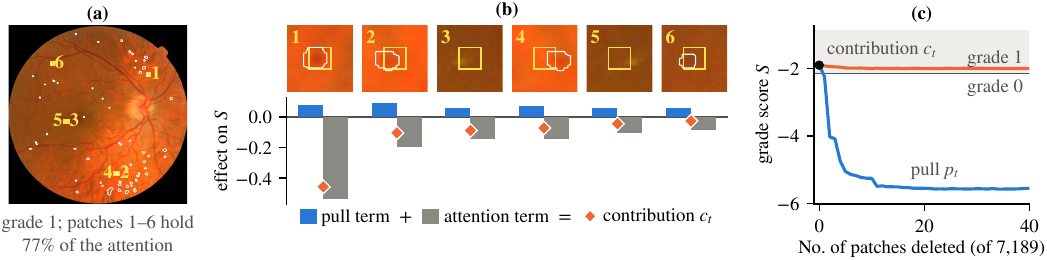}
\caption{The contribution $c_t$ contains the causal pull $p_t$, yet can rank lesions last. By Eq.~\ref{eq:identity}, $c_t$ sums a pull term and an attention term. The pull term sums to zero in that head, so completeness rests on the attention term. The grade score $S$ is represented by $S_\varnothing + \sum_{h,t} c_{h,t}$. In an example image (predicted as grade~1), when $S < S_\varnothing$, the attention term inverts the attention map and $c_t$ ranks the lesions of interest last. (a) Expert lesions (white annotations); top-6 attended patches are highlighted by yellow boxes. (b) Both terms per patch, summed over 12 heads. The negative attention term (grey bar) outweighs the positive pull term (blue bar), so $c_t$ (diamonds) ranks these 6 patches last of 7,189 patches and $p_t$ ranks them correctly. (c) How DR grade score $S$ varies as top-ranked patches are dropped; grey line: DR grade threshold between grade 1 and grade 0. Removing one $p_t$ patch lowers the grade, while removing 40 $c_t$ patches does not.}
\label{fig:main}
\end{figure*}

Deep learning models for diabetic retinopathy (DR) classification achieve specialist-level accuracy~\cite{gulshan2016}, but clinical adoption is limited by a lack of causal and pathology-grounded explanations that operate efficiently at high resolution. A recent benchmark of six post-hoc methods on a DR vision transformer reported near-zero faithfulness for all the evaluated approaches~\cite{cascone2026benchmark}.

Attention-pooling heads, including the multi-head attention-pooling (MAP) head of SigLIP~2~\cite{lee2019settransformer,tschannen2025siglip2} and attention-based multiple instance learning (MIL), provide attribution maps via a single forward pass, leveraging attention coefficients and fixed-attention decompositions such as Value-Zeroing~\cite{mohebbi2023valuezeroing}. Decompositions that propagate through the non-linear head are either first-order approximations~\cite{gandelsman2024clip} or require retraining the model~\cite{javed2022additive}. The causal validity of attention as an explanation remains contested~\cite{jain2019attention}, and in MIL settings, attention frequently fails removal-based evaluation~\cite{hense2024xmil}. Attribution based on token removal requires the explanation to predict the change in grade upon patch ablation. Fixed-attention decompositions can accumulate evidence to reflect the original prediction but do not capture the causal effect of removing a patch (Fig.~\ref{fig:main}). We validate these theoretical insights empirically on a NaFlex-ViT with a MAP head, making the following key contributions:
\begin{enumerate}
\item \textbf{Exact decomposition through a non-linear head.} An integrated-gradient read-out decomposes the DR grade score $S$ exactly into the contribution $c_t$, and yields the pull $p_t$, the closed-form effect of removing a patch and renormalizing attention in the pool.
\item \textbf{An identity linking attribution and causality.} The two maps satisfy
$c_{h,t}=(1-a_{h,t})\,p_{h,t}+a_{h,t}E_h.$ Since the pull term \((1-a_{h,t})\,p_{h,t}\) sums to zero over all patches. By the completeness axiom~\cite{sundararajan2017ig}, the total contribution is therefore exactly captured by the \(a_{h,t}E_h\) term.
\item \textbf{Causal maps, at a fraction of the cost.} The pull $p_{t}$ and the attention weights $\bar a_t$ beat Attention Rollout, Grad-CAM, AttnLRP and occlusion in 11 of 12 comparisons, $800\times$ faster than occlusion, and localize lesions out-of-distribution at 70\% of the average precision of a supervised segmentation model.
\end{enumerate}

% new methodology
\section{Methodology}
\label{sec:method}

\noindent\textbf{Attention pooling.}
Given $N$ patch tokens $\{\mathbf{x}_t\}_{t=1}^{N}$ from a vision transformer backbone, a multi-head attention-pooling (MAP) head~\cite{lee2019settransformer} with learned queries $\mathbf{q}_h$ computes in each head $h$ the softmax attention weights $a_{h,t}$, the values $\mathbf{v}_{h,t}=\mathbf{W}_V^h\mathbf{x}_t+\mathbf{b}_V^h$, the head vectors $\mathbf{z}_h=\sum_t a_{h,t}\mathbf{v}_{h,t}$ and the pooled representation $\mathbf{z} = \mathbf{b}_O + \sum_{h=1}^{H} \mathbf{W}_O^h \mathbf{z}_h$. A nonlinear function $\varphi$, made of a LayerNorm, a residual MLP followed by a linear layer, maps $\mathbf{z}$ to one ordinal grade score $S = \varphi(\mathbf{z})$, and the grade is the number of validation-recalibrated thresholds that $S$ exceeds. The mean attention map $\bar a_t=\sum_h a_{h,t}/H$ is an unsigned attribution map that cannot tell how each patch contributes to the final score $S$. \\
\noindent\textbf{A signed decomposition of the score.}
An attribution is complete~\cite{sundararajan2017ig} if its patch terms sum exactly to $S - S_{ref}$. Our reference is the empty pool, $\mathbf{v}_{h,t}=\mathbf{0}$, which gives $\mathbf{z}=\mathbf{b}_O$ and the constant score $S_\varnothing=\varphi(\mathbf{b}_O)$. Integrating the gradient of $\varphi$ along the straight path $\mathbf{z}_\alpha=\mathbf{b}_O+\alpha(\mathbf{z}-\mathbf{b}_O)$ makes the head exactly linear for this image, $S-S_\varnothing=\bar{\mathbf{g}}^\top(\mathbf{z}-\mathbf{b}_O)$, with
\begin{equation}
\bar{\mathbf{g}} = \int_0^1 \nabla_{\mathbf{z}} \varphi(\mathbf{z}_\alpha)\,d\alpha \approx \tfrac{1}{64}\textstyle\sum_{k=1}^{64} \nabla_{\mathbf{z}}\varphi\bigl(\mathbf{z}_{(k-1/2)/64}\bigr)
\label{eq:ig}
\end{equation}
We can substitute for $\mathbf{z}$ and simplify with $\mathbf{r}_h = \mathbf{W}^{h \top}_O \bar{\mathbf{g}}$ and patch value $e_{h,t} = \mathbf{r}_h^\top \mathbf{v}_{h,t}$ and the head evidence $E_h = \sum_t a_{h,t}e_{h,t} = \mathbf{r}_h^\top \mathbf{z}_h$, therefore the completeness axiom~\cite{sundararajan2017ig} holds:
\begin{equation}
\textstyle S = S_\varnothing + \sum_{h} E_h = S_\varnothing + \sum_{h,t} c_{h,t}, \quad c_{h,t} = a_{h,t} e_{h,t},
\label{eq:complete}
\end{equation}
where $c_{h,t}$ is the \emph{contribution} of patch $t$ through head $h$, with map $c_t=\sum_h c_{h,t}$.

\noindent\textbf{Accumulation versus removal: The pull.}
The contribution answers whether the evidence $E_h$ accumulates toward the score $S$. Methods like Value-Zeroing~\cite{mohebbi2023valuezeroing} set a patch's value to zero and measure the resulting change in the output, but this does not accurately capture the observed behavior; without a patch, attention would shift to the other patches. Therefore, we need to renormalize attention $a_{h,s}\to a_{h,s}/(1-a_{h,t})$. Removing patch $t$ and renormalizing gives $\mathbf{z}_h'$ in closed-form, the single-token case of the attention-deletion response of~\cite{huang2026rope}. We remove patch $t$ and obtain the renormalized vector $\mathbf{z}_h'$. Read out with $\mathbf{r}_h$, this difference between the perturbed vector $\mathbf{z}_h'$ and $\mathbf{z}_h$ defines the \emph{pull} of patch $t$ in head $h$.
\begin{equation}
p_{h,t} = \mathbf{r}_h^\top\bigl(\mathbf{z}_h - \mathbf{z}_h'\bigr) = \frac{a_{h,t}}{1 - a_{h,t}}\bigl(e_{h,t} - E_h\bigr),
\label{eq:pull}
\end{equation}
with map $p_t=\sum_h p_{h,t}$. We reuse $\bar{\mathbf{g}}$ from Eq.~\eqref{eq:ig} to approximate the readout $\mathbf{r}_h$, which predicts the exact score shift with high correlation, as shown in Fig.~\ref{fig:pool}a. 

\noindent\textbf{The identity and the completeness axiom.} Substituting Eq.~\eqref{eq:pull} with $c_{h,t}=a_{h,t}e_{h,t}$ relates the two maps:
\begin{equation}
c_{h,t} = (1 - a_{h,t})\, p_{h,t} + a_{h,t} E_h, \quad \textstyle\sum_t (1 - a_{h,t})\, p_{h,t} = 0.
\label{eq:identity}
\end{equation}
The pull term $(1-a_{h,t})p_{h,t}$ carries the removal effect but sums to zero in each head, so completeness rests entirely on the attention term $a_{h,t}E_h$, which is the attention scaled by the net head evidence $E_h$, and this term is \emph{inverted} when $E_h<0$, as is common in low-grade images. A patch with $E_h<e_{h,t}<0$ then raises $S$ ($p_{h,t}>0$) yet receives $c_{h,t}<0$, even when the head attends mostly to lesions (Fig.~\ref{fig:main}). The \emph{alignment} $q=\operatorname{corr}_t\bigl(\sum_h a_{h,t}E_h,\,\bar a_t\bigr)$, the Pearson correlation over patches, separates the two cases on each image. The identity predicts that $c_t$ departs from removal when $q<0$, while $p_t$, which lacks the attention term, does not depend on $q$ as shown in Fig.~\ref{fig:pool}a.

\section{Experiments and Results}
\label{sec:results}

\begin{table*}[t]
\centering
\caption{Causal tests (token dropping) and lesion $AP$ on MAPLES-DR, three seeds. Brackets: 95\% CIs of the causal tests; $^{\dagger}$: significantly better than all four baselines ($p_{\mathrm{Holm}}<0.05$, 12 comparisons per map). Best attribution map in bold red, second best underlined in blue. Cost per image at 1,536~px; (--): not applicable.}
\label{tab:main}
\small
\renewcommand{\arraystretch}{1.0}
\setlength{\tabcolsep}{4.5pt}
\newcommand{\ci}[1]{\textcolor{black!55}{[#1]}}
\newcommand{\sig}{\rlap{$^{\dagger}$}}
\begin{tabular*}{\textwidth}{@{}l@{\extracolsep{\fill}} r@{\extracolsep{0pt}\hspace{7pt}}l@{\extracolsep{\fill}} r@{\extracolsep{0pt}\hspace{7pt}}l@{\extracolsep{\fill}} r@{\extracolsep{0pt}\hspace{7pt}}l@{\extracolsep{\fill}} cc rr@{}}
\toprule
& \multicolumn{6}{c}{Causal tests (MAPLES-DR)} & \multicolumn{2}{c}{Lesion $AP$$\uparrow$} & \multicolumn{2}{c}{Cost per image}\\
\cmidrule(lr){2-7}\cmidrule(lr){8-9}\cmidrule(l){10-11}
Map & \multicolumn{2}{c}{$N$ (\%)$\downarrow$} & \multicolumn{2}{c}{$Su$ (\%)$\downarrow$} & \multicolumn{2}{c}{$FC$$\uparrow$} & MAPLES & DDR & TFLOPs & Time (s)\\
\midrule
\multicolumn{11}{@{}l}{\textbf{Attention Decomposition (Ours)}}\\
\cdashline{1-11}\noalign{\vskip 2pt}
\quad Attention weights $\bar a_t$ & \best{0.61}\sig & \ci{0.50, 0.76} & 3.9 & \ci{3.0, 5.2} & \second{0.47}\sig & \ci{0.44, 0.50} & \second{0.275} & \best{0.274} & \best{4.7} & \best{0.14}\\
\quad Contribution $c_t$ & 0.96 & \ci{0.61, 1.74} & 3.5 & \ci{2.6, 4.8} & 0.26 & \ci{0.01, 0.42} & 0.244 & \second{0.263} & \best{4.7} & \second{0.23}\\
\quad Pull $p_t$ & \second{0.71}\sig & \ci{0.57, 0.89} & \second{3.0} & \ci{2.2, 4.6} & \best{0.53}\sig & \ci{0.49, 0.57} & \best{0.277} & 0.252 & \best{4.7} & 0.27\\
\midrule
\multicolumn{11}{@{}l}{\textbf{Baselines \& Controls}}\\
\cdashline{1-11}\noalign{\vskip 2pt}
\quad Rollout~\cite{abnar2020rollout} & 1.04 & \ci{0.73, 1.46} & 8.1 & \ci{6.4, 10.0} & 0.17 & \ci{0.12, 0.21} & 0.195 & 0.247 & \second{6.3} & 2.3\\
\quad Grad-CAM~\cite{selvaraju2017gradcam} & 5.3 & \ci{1.5, 36.4} & 8.3 & \ci{4.6, 19.2} & 0.22 & \ci{0.09, 0.31} & 0.124 & 0.138 & 14.2 & 2.6\\
\quad AttnLRP~\cite{achtibat2024attnlrp} & 6.2 & \ci{3.5, 11.6} & 7.7 & \ci{5.9, 10.0} & $-$0.15 & \ci{$-$0.22, $-$0.06} & 0.163 & 0.206 & 25.3 & 4.9\\
\quad Occlusion~\cite{zeiler2014visualizing} & 2.35 & \ci{1.64, 3.37} & \best{1.6} & \ci{1.1, 2.7} & 0.32 & \ci{0.28, 0.36} & 0.112 & 0.107 & 8,391 & 227\\
\cdashline{1-11}\noalign{\vskip 2pt}
\multicolumn{11}{@{}l}{\textit{Controls}}\\
\quad Expert lesion masks & 1.35 & \ci{1.03, 1.79} & 2.7 & \ci{2.1, 3.8} & 0.30 & \ci{0.26, 0.34} & -- & -- & -- & --\\
\quad Random order & 71 & \ci{64, 78} & 12 & \ci{8.4, 16.9} & 0.00 & \ci{$-$0.01, 0.02} & -- & -- & -- & --\\
\midrule
\multicolumn{11}{@{}l}{\textbf{SOTA Segmentation Models}}\\
\cdashline{1-11}\noalign{\vskip 2pt}
\quad HRDecoder~\cite{ding2024hrdecoder} & -- & & -- & & -- & & 0.396 & 0.639 & -- & --\\
\quad nnU-Net~\cite{isensee2021nnunet} & -- & & -- & & -- & & 0.384 & 0.582 & -- & --\\
\bottomrule
\end{tabular*}
\end{table*}

\noindent\textbf{Data.} We train on EyePACS~\cite{eyepacs2015}, BRSET~\cite{nakayama2024brset} and APTOS~2019~\cite{aptos2019}, split into $34,032$ training, $3,813$ validation and $3,267$ test images. We run all causality and localization experiments out-of-distribution (OOD) on the MAPLES-DR~\cite{lepetit2024maples} dataset with expert lesion masks, and additionally test on the DDR~\cite{li2019ddr} dataset only for lesion localization. We tune all baseline map configurations on the held-out IDRiD dataset~\cite{porwal2018idrid}.

\noindent\textbf{Model and training.} We use NaFlexViT-B/16 with 93\,M parameters and SigLIP~2 weights~\cite{tschannen2025siglip2}. This variant supports variable-length token input, which is useful for dropping a token at the image level. We fine-tune the model with the CORAL objective \cite{cao2020coral} using three seeds in two stages: 12 epochs at 1,024~px, then 2 epochs at 1,536~px. The trained classifier reaches a quadratic weighted kappa (QWK) of $0.836\pm0.013$ and a referable-DR AUC of $0.972\pm0.002$ on the OOD Messidor-2 dataset~\cite{krause2018}. All experiments ran on one 96-GB NVIDIA RTX PRO 6000.

\noindent\textbf{Baselines and protocol.} We compare our attribution maps against attention rollout from the pool query~\cite{abnar2020rollout}, Grad-CAM at block 8~\cite{selvaraju2017gradcam}, AttnLRP with the gamma and CP-LRP rules~\cite{achtibat2024attnlrp,ali2022cplrp} and $2\times2$-patch occlusion~\cite{zeiler2014visualizing}; expert lesion masks and random rankings serve as controls. We remove patches via token dropping, which avoids pixel-masking artifacts~\cite{jain2022missingness}. Over top-ranked percentages of patches $f\in\{0.25,\dots,50\}\%$, \emph{Necessity} ($N$) is the smallest $f$ whose deletion lowers the predicted grade, and \emph{Sufficiency} ($Su$) is the smallest $f$ whose insertion restores it~\cite{petsiuk2018rise}, starting from an empty fundus that keeps only the tokens outside the field of view. Both metrics are geometric means computed over the $143$ images that all three seeds grade $\geq1$, with $f=100\%$ when the grade never changes. \emph{Faithfulness Correlation} ($FC$)~\cite{bhatt2020evaluating} tests the values of a map, not only its ranking: for each image, we remove 128 random groups of $144$ patches, each covering $2\%$ of the fundus, and $FC$ is the Pearson correlation between the map summed over a group and the drop $\Delta S$ that its removal causes. Significance is assessed with a paired hierarchical bootstrap over test images and seeds ($5,000$ resamples) and Holm-corrected over the $12$ baseline comparisons; details are in the code.

\begin{figure}[t]
\centering
\includegraphics[width=\columnwidth]{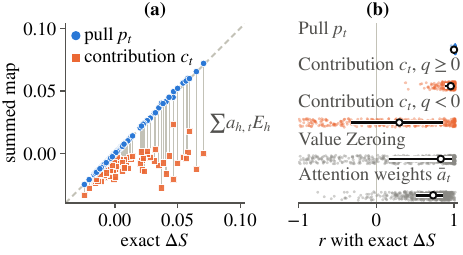}
\caption{The pull $p_t$ predicts removal from the pool. We remove random patch groups from the pooling head (attention logits set to $-\infty$ in all heads, backbone fixed) and record the exact drop $\Delta S$ in grade score. (a) Each point is one of 128 groups covering 2\% of the fundus, plotted as exact $\Delta S$ ($x$) against the map summed over the group ($y$). The pull $p_t$ lies on the identity line; the contribution falls below it. The grey line equals the summed attention term $\sum a_{h,t}E_h$ of Eq.~\eqref{eq:identity}. (b) Pearson $r$ with $\Delta S$ for each map, one dot per image and seed; $c_t$ is split by the sign of $q$. Circles: median; bars: interquartile range.}
\label{fig:pool}
\end{figure}

\noindent\textbf{Closed-form causality, and why fixed-attention maps fail.} Removing each random group from the pool (logits $-\infty$ in all heads), the summed pull $p_t$ tracks the true change of $S$ with median Pearson $r=1.000$ on every seed, and the contribution $c_t$ with only $r=0.854\,[0.661, 0.922]$ (Fig.~\ref{fig:pool}). For such small groups, $c_t$ is off by the attention term $\sum_h a_{h,t}E_h$ (Fig.~\ref{fig:pool}a), so it fails when $q<0$: median $r=0.30$ against $0.96$ for $q\ge0$ (Fig.~\ref{fig:pool}b). In end-to-end testing under token dropping (Table~\ref{tab:main}), $p_t$ is the most faithful map with an $FC$ of $0.53$, whereas $c_t$ beats the baselines in $4$ of its $12$ comparisons, with an $FC$ of $0.26$. Across images, $c_t$'s FC rises with $q$, whereas $p_t$ does not depend on $q$ ($FC$ ranges from $0.49-0.57$).

\begin{table}[t]
\centering
\caption{CORAL against a cross-entropy (CE) head, MAPLES-DR, three seeds. $^{\dagger}$: significant difference between the heads (bootstrap, Holm correction over the six causal tests). Best in bold red, second underlined in blue.}
\label{tab:ce}
\small
\renewcommand{\arraystretch}{1.0}
\setlength{\tabcolsep}{4pt}
\begin{tabular*}{\columnwidth}{@{}l@{\extracolsep{\fill}}lcccc@{}}
\toprule
Map & Loss & $N$ (\%)$\downarrow$ & $Su$ (\%)$\downarrow$ & $FC$$\uparrow$ & $AP$$\uparrow$\\
\midrule
Attention & CORAL~\cite{cao2020coral} & \best{0.61} & 3.9 & \second{0.47}\rlap{$^{\dagger}$} & \second{0.275}\\
weights $\bar a_t$ & CE & 1.10 & 3.9 & 0.29 & 0.211\\
\midrule
Pull $p_t$ & CORAL~\cite{cao2020coral} & \second{0.71} & \second{3.0} & \best{0.53} & \best{0.277}\\
& CE & 0.76 & \best{2.6} & 0.45 & 0.269\\
\bottomrule
\end{tabular*}
\end{table}

\noindent\textbf{Grade-level causal tests.} The attention weights $\bar a_t$ and the pull $p_t$ each beat every baseline in $11$ of $12$ Holm-corrected comparisons ($p_{\mathrm{Holm}}\le0.018$ and $\le0.032$; Table~\ref{tab:main}). $\bar a_t$ needs the fewest patches to lower the grade ($N=0.61\%$; deleting its top $1\%$ lowers the grade in $74\%$ of the images, against $71\%$ for $p_t$), and $p_t$ has the highest $FC$, $0.531$, above $\bar a_t$ by $0.066\,[0.032, 0.101]$ and $c_t$ by $0.276\,[0.106, 0.496]$. Occlusion has better sufficiency ($1.6\%$ against $3.9\%$ and $3.0\%$), but our maps are about $800\times$ faster and use $1,800\times$ fewer FLOPs.

The attention weights $\bar a_t$ have an edge over $p_t$ on \emph{Necessity} ($0.61\%$ vs $0.71\%$, $p>0.05$ (n.s.)), since $p_{h,t}$ scales with $a_{h,t}$ (Eq.~\eqref{eq:pull}) but, being non-negative, $\bar a_t$ cannot indicate whether removing a patch raises or lowers $S$. By contrast, $p_t$ gives the signed, first-order effect of removing each patch from the pool, and leads on $FC$ ($0.531$ vs $0.465$). The pull adds the sign that attention lacks.

\noindent\textbf{Lesions without pixel labels.} We score each upsampled map by its per-image average precision ($AP$) on the binary union of lesion masks. On MAPLES-DR, $p_t$ and $\bar a_t$ have the highest $AP$ ($0.277$ and $0.275$); $p_t$ exceeds $c_t$ by $0.033$ ($p<0.05$) and Rollout, AttnLRP, Grad-CAM and Occlusion by $0.08$--$0.17$ ($p<0.001$). It reaches $70\%$ of the $AP$ of HRDecoder~\cite{ding2024hrdecoder} and matches its pixel $AUC$ ($0.847$ against $0.846$), although the segmentation models are trained on the lesion masks of DDR, IDRiD and e-ophtha~\cite{decenciere2013teleophta}. On DDR, inside that training distribution, $p_t$ performs like $\bar a_t$, $c_t$ and Rollout, and better than AttnLRP, Grad-CAM and Occlusion ($p\le0.016$).

\noindent\textbf{The ordinal head shapes the attention.} Replacing CORAL with a 5-class cross-entropy loss, with the expected grade as the score and everything else identical, gives a $QWK$ of $0.822$ vs.\ $0.829$ in distribution and $0.866$ vs.\ $0.836$ on the OOD Messidor-2, but $\bar a_t$ becomes less faithful ($FC$ $0.47\to0.29$, $p_{\mathrm{Holm}}<0.001$; Table~\ref{tab:ce}) and localizes lesions worse ($AP$ $0.275\to0.211$, $p<0.05$). The pull $p_t$ is as causal for either head ($p_{\mathrm{Holm}}\ge0.21$), since Eqs.~\eqref{eq:complete}--\eqref{eq:identity} hold for any scalar head: the ordinal head points attention to patches whose removal lowers the grade.

\section{Conclusion}
\label{sec:conclusion}

The three maps answer three questions. The attention weights $\bar a_t$ show where the evidence is, at no cost beyond the forward pass; the pull $p_t$ adds its sign and the effect of its removal; the contribution $c_t$ accumulates to the score $S$. Value-Zeroing and every decomposition that holds the attention fixed, including Additive MIL~\cite{javed2022additive}, answer the accumulation question. Since the 12 head evidences have mixed signs in $62 - 74\%$ of test images across seeds, we recommend reporting the alignment $q$ before reading such maps as causal: it is free, and a negative $q$ flags them as unreliable, so that $p_t$ should be shown instead. An attention-pooling head thus explains its own grade in one forward pass. Our study is limited to one architecture with three seeds; $p_t$ is an approximation at the pool, and occlusion retains better sufficiency. The end-to-end tests use a backbone trained on variable-length token sequences; on fixed-length vision transformers, token dropping may confound the removal of evidence with a distribution shift~\cite{jain2022missingness}, although Eqs.~\eqref{eq:complete}--\eqref{eq:identity} and the pool-level $FC$ do not depend on this. Testing such backbones and explaining grade~0, where deletion cannot lower the score, remain open questions.

\section{Compliance with Ethical Standards}
\label{sec:ethics}
This research study was conducted retrospectively using human subject data made available in open access (EyePACS, BRSET, APTOS~2019, Messidor-2, IDRiD, MAPLES-DR and DDR). Ethical approval was not required, as confirmed by the licenses attached to the open-access data.

\bibliographystyle{IEEEbib}
\bibliography{refs}

\end{document}